\documentclass[10pt,twocolumn,letterpaper]{article}

\usepackage[pagenumbers]{cvpr} 

\usepackage{lineno}
\usepackage{wrapfig}
\usepackage{multirow}
\usepackage{graphicx}
\usepackage{amsmath}
\usepackage{amssymb}
\usepackage{booktabs}

\usepackage{array} 
\usepackage{xcolor}
\usepackage{courier}
\usepackage{listings}

\usepackage{bm}
\usepackage{bbding}
\usepackage{color}
\usepackage{pifont}
\usepackage{csquotes}

\usepackage{algorithmic}
\usepackage{algorithm}
\usepackage{gensymb}

\definecolor{cvprblue}{rgb}{0.21,0.49,0.74}
\usepackage[pagebackref,breaklinks,colorlinks,allcolors=cvprblue]{hyperref}

\def\paperID{9375} 
\def\confName{CVPR}
\def\confYear{2025}

\title{
EchoWorld: 
Learning Motion-Aware World Models \\
for Echocardiography Probe Guidance
}

\author{
Yang Yue$^1$\thanks{Equal contribution.\ \ \ \ \ \ \ \ \ \ \ \ \ \ \ \textsuperscript{\Envelope}Corresponding authors.} \ \ \ \ \ 
Yulin Wang$^1$$^*$ \ \ \ \ 
Haojun Jiang$^1$ \ \ \ \  
Pan Liu$^2$ \ \ \ \ 
Shiji Song$^1$ \ \ \ \ 
Gao Huang$^{1}$ \!\textsuperscript{\Envelope}\\
{\small $^1$Tsinghua University\ \ \ \ \ \ $^2$PLA General Hospital}\\[-0.5ex]
{\small\texttt{yueyang22@mails.tsinghua.edu.cn,\ gaohuang@tsinghua.edu.cn}}
}

\newcommand{\methodname}{EchoWorld}
\begin{document}
\maketitle

\begin{abstract}


Echocardiography is crucial for cardiovascular disease detection but relies heavily on experienced sonographers. Echocardiography probe guidance systems, which provide real-time movement instructions for acquiring standard plane images, offer a promising solution for AI-assisted or fully autonomous scanning. However, developing effective machine learning models for this task remains challenging, as they must grasp heart anatomy and the intricate interplay between probe motion and visual signals.
To address this, we present \methodname{}, a motion-aware world modeling framework for probe guidance that encodes anatomical knowledge and motion-induced visual dynamics, while effectively leveraging past visual-motion sequences to enhance guidance precision.
\methodname{} employs a pre-training strategy inspired by world modeling principles, where the model predicts masked anatomical regions and simulates the visual outcomes of probe adjustments. Built upon this pre-trained model, we introduce a motion-aware attention mechanism in the fine-tuning stage that effectively integrates historical visual-motion data, enabling precise and adaptive probe guidance.
Trained on more than one million ultrasound images from over 200 routine scans, \methodname{} effectively captures key echocardiographic knowledge, as validated by qualitative analysis. Moreover, our method significantly reduces guidance errors compared to existing visual backbones and guidance frameworks, excelling in both single-frame and sequential evaluation protocols. Code is available at \url{https://github.com/LeapLabTHU/EchoWorld}.


\end{abstract}    

\begin{figure*}[h]
    \centering
    \includegraphics[width=\linewidth]{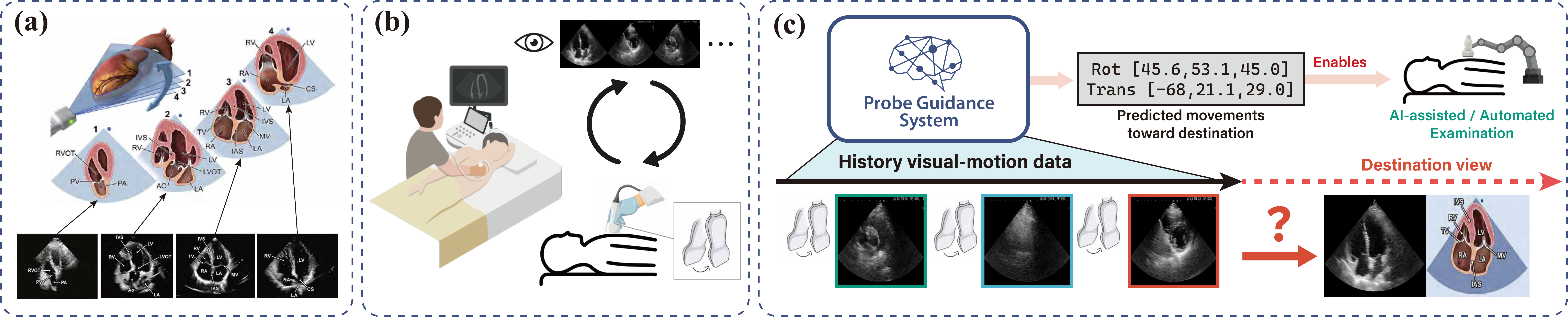}
    \vspace{-20pt}
    \caption{\textbf{Overview of cardiac ultrasound and the probe guidance task.} (a) The ultrasound probe captures cross-sectional views of the heart, with variations in probe position and orientation corresponding to different anatomical structures. 
    (b) During the ultrasound scanning process, the sonographer maneuvers the probe on the patient's chest, continuously adjusting its position and orientation based on real-time visual feedback.
    (c) A probe guidance system can potentially automate the scanning process by predicting the necessary probe movements to reach a target view, utilizing historical visual-motion data.
    }
    \label{fig:ultrasound}
    \vspace{-5pt}
\end{figure*}

\begin{figure*}[h]
    \centering
    \includegraphics[width=\linewidth]{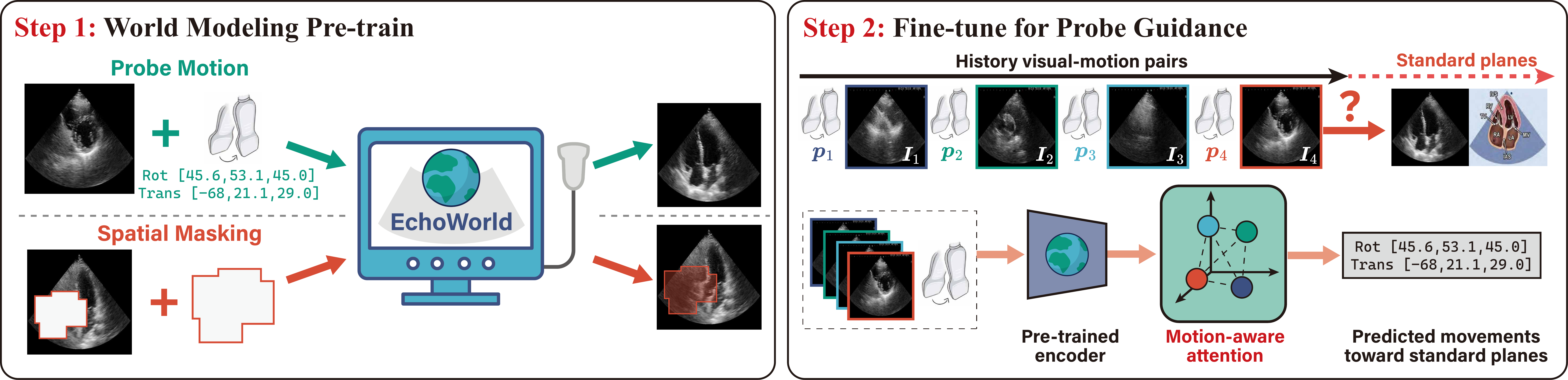}
    \vspace{-20pt}
    \caption{\textbf{Overview of the proposed framework.} Left: We pre-train a cardiac world model to capture ultrasound knowledge through spatial and motion modeling tasks. Right: The pre-trained model is fine-tuned for probe guidance, incorporating a motion-aware attention mechanism to effectively integrate visual-motion features.}
    \label{fig:idea}
    \vspace{-15pt}
\end{figure*}

\section{Introduction}
\label{sec:intro}

Cardiovascular disease remains one of the leading causes of death worldwide \cite{roth2017global,song2020global}, making timely and accurate diagnosis critical to saving lives. 
Among the various diagnostic tools available, echocardiography stands out as a non-invasive, cost-effective, and widely accessible method for assessing cardiac health. In this method, a \textit{probe} emits high-frequency sound waves into the body, which are reflected by heart structures and captured to generate real-time images (Figure \ref{fig:ultrasound}(a)).
However, performing cardiac ultrasound scans requires the sonographer to carefully maneuver the probe to acquire key sectional views of the heart (Figure \ref{fig:ultrasound}(a-b)), a task that demands extensive anatomical knowledge and experience. This complexity, coupled with a global shortage of qualified sonographers, limits the accessibility of ultrasound services, particularly in less developed regions. This challenge has motivated the development of probe guidance systems \cite{droste2020automatic,narang2021utility, sabo2023real, jiang2024cardiac} capable of assisting less experienced sonographers or, in the longer term, enabling fully autonomous ultrasound scanning robots.
Such systems hold the potential to democratize cardiac care by providing real-time, actionable feedback, which can significantly improve the efficiency of the scanning process.

As shown in Figure \ref{fig:ultrasound}(c), probe guidance in echocardiography can be formulated as a vision-based sequential prediction problem, where a model needs to predict the necessary probe movement vectors to reach each target view, utilizing historical visual-motion data.
However, unlike other computer vision tasks in medical imaging, it presents the unique challenge of integrating motion data with dynamic visual observations. 
Autonomous systems must not only understand the complex anatomical structures of the heart (\emph{e.g.} chambers, valves, and vessels) but also how these structures are represented in ultrasound images as the probe moves and changes position.
While early efforts in probe guidance \cite{droste2020automatic, narang2021utility, shida2023automated} have made some progress in developing assistive and autonomous scanning systems, few studies focus on a fundamental problem: \emph{How can we develop a principled approach that effectively learns essential medical knowledge while seamlessly integrating visual and motion data for precise probe guidance?}


In this paper, we present \methodname{}, a motion-aware world modeling framework that begins by pre-training a strong representation model on visual-motion data, followed by fine-tuning the model with a novel motion-aware attention mechanism that allows seamless integration of motion information with visual features.

The first stage, world model pre-training, is designed to encode rich, common-sense knowledge about the world \cite{ha2018world,lecun2022path}, which can potentially capture the heart’s anatomical structure and the relationships between different probe positions, allowing the system to guide a sonographer much like an experienced driver navigating through the city with an internalized map. As demonstrated in the left part of Figure \ref{fig:idea}, our cardiac world model encodes two key dimensions of echocardiology knowledge: 1) the appearance of anatomical structures (\emph{e.g.}, ventricle, valves, and septums) in cardiac ultrasound images and 2) the changing dynamic of visual signals following the probe motions.

Building on this pre-trained world model, we introduce a guidance prediction module with a motion-aware attention mechanism that integrates historical image-pose data, as shown in the right part of Figure \ref{fig:idea}. Unlike existing methods that typically organize the data into interleaved visual-action sequences, our motion-aware attention mechanism embeds 3D relative pose differences into the attention features.  This enables motion-aware interactions across image frames, allowing the system to better track anatomical structures and produce more accurate predictions.

Our proposed framework, \methodname, is built upon a cardiac ultrasound scanning dataset derived from routine clinical examinations. We empirically show that \methodname{} can act as a cardiac ultrasound simulator enriched with anatomical knowledge. Additionally, we compare \methodname{} against a wide range of pre-trained models and existing probe guidance methods. Our model consistently outperforms these approaches in acquiring ten standard planes, achieving lower guidance errors across both single-frame and sequential evaluation protocols. 
Analytical results and ablation studies further validate the effectiveness of the proposed framework. 



\section{Related Work}
\label{sec:related}
\textbf{World Models.}
The concept of the world model was first introduced in psychology \cite{craik1967nature} and later adapted for model-predictive control \cite{bryson2018applied,camacho2007constrained} and reinforcement learning \cite{ha2018recurrent,ha2018world,hafner2019dream,hafner2020mastering}. In these contexts, a world model typically predicts future states of the environment based on an agent’s actions. More recently, the development of general-purpose world models that encompass broad, commonsense understanding has been recognized as a key step toward achieving general artificial intelligence \cite{lecun2022path,runwayml2023worldmodels}. The advent of large-scale video generation models has further highlighted this potential, demonstrating their ability to serve as physical simulators \cite{openai2024videogeneration,kang2024far}, driving scene simulators \cite{hu2023gaia,gao2024vista,yang2024generalized}, and game engines \cite{valevski2024diffusion,oasis,alonso2024diffusion}. 
World modeling has also proven to be a powerful tool for representation learning, producing structured and informative representations that capture complex world dynamics with minimal supervision \cite{hafner2019dream, hafner2020mastering, hafner2023mastering,ijepa,bardes2023v,iwm,baevski2023efficient,baevski2022data2vec}.
Our proposed framework utilizes world modeling as a pretext task to develop a robust visual representation model. Our analytical experiments show that \methodname{} effectively captures echocardiography knowledge for probe guidance. When augmented with a diffusion model, it can potentially function as a simulator for free-hand cardiac ultrasound scanning, capable of predicting visual changes based on probe movements.

\textbf{AI for ultrasound.} Recent AI advances have driven significant progress in ultrasound applications, advancing tasks like segmentation \cite{stojanovski2023echo, deng2024memsam, lin2024beyond}, 3D reconstruction \cite{yan2024fine}, and diagnostic support \cite{ouyang2020video,wang2024thyroid}. 
Recent efforts also focus on creating foundation models for ultrasound, as exemplified by USFM \cite{jiao2024usfm}, which explores self-supervised learning, and EchoCLIP \cite{christensen2024vision}, which utilizes multimodal learning techniques. Another critical AI application in ultrasound is probe guidance, aimed at assisting novices and inexperienced sonographers \cite{droste2020automatic, narang2021utility, sabo2023real} or enabling fully autonomous robotic scanning \cite{shida2023automated, shida2023diagnostic}. For instance, US-GuideNet \cite{droste2020automatic} provides rotational guidance for free-hand obstetric ultrasound, while  \citet{shida2023automated,shida2023diagnostic} investigates search-based methods to acquire the Parasternal Long-Axis (PLAX) plane in cardiac ultrasound. 
Current ultrasound probe guidance methods primarily leverage imitation learning \cite{droste2020automatic, jiang2024cardiac, jiang2024sequence, men2023gaze} or reinforcement learning \cite{amadou2024goal, li2023rl}.
The latter typically relies on CT-derived simulations
, whereas imitation learning---the approach adopted in our study---directly learns from expert demonstrations, presenting a scalable approach \cite{jiang2024cardiac} in line with advances in general-purpose robotic control \cite{brohan2023rt, kim2024openvla,o2023open}. 
While prior research has focused on probe control for ultrasound scanning, 
little attention has been given to
representation learning strategies and network architectures for ultrasound data.
In this paper, we seek to bridge these gaps by proposing a motion-aware world modeling framework tailored for ultrasound.









\section{Background and Notations}

Before presenting our method, we briefly overview the technical background relevant to the probe guidance task and the format of the data employed in our study.

\subsection{Cardiac Ultrasound}
Cardiac ultrasound, or echocardiography, involves the use of a transducer (or \textit{probe}) that emits high-frequency sound waves into the body, which are then reflected by heart structures and captured to generate real-time images. 
These images depict two-dimensional cross-sectional views, or \textit{planes}, of the heart, showcasing its chambers, walls, valves, and blood flow dynamics, and are essential for diagnosing a variety of cardiac conditions. The specific position, orientation, and tilt of the probe determine the captured plane, as shown in Figure \ref{fig:ultrasound}(a). 
A \textit{standard plane} in cardiac ultrasound is a predefined cross-sectional view, systematically used to assess specific heart structures. For instance, the Parasternal Short-Axis (PSAX) plane offers a horizontal slice through the heart, which is instrumental in evaluating heart muscle thickness and valve function. Obtaining these views requires precise probe maneuvers; the sonographer skillfully maneuvers the probe on the patient's chest, carefully adjusting its position, angle, and pressure to capture the desired two-dimensional \enquote{slice} of the heart. 

\begin{figure}[t]
    \centering
\includegraphics[width=\linewidth]{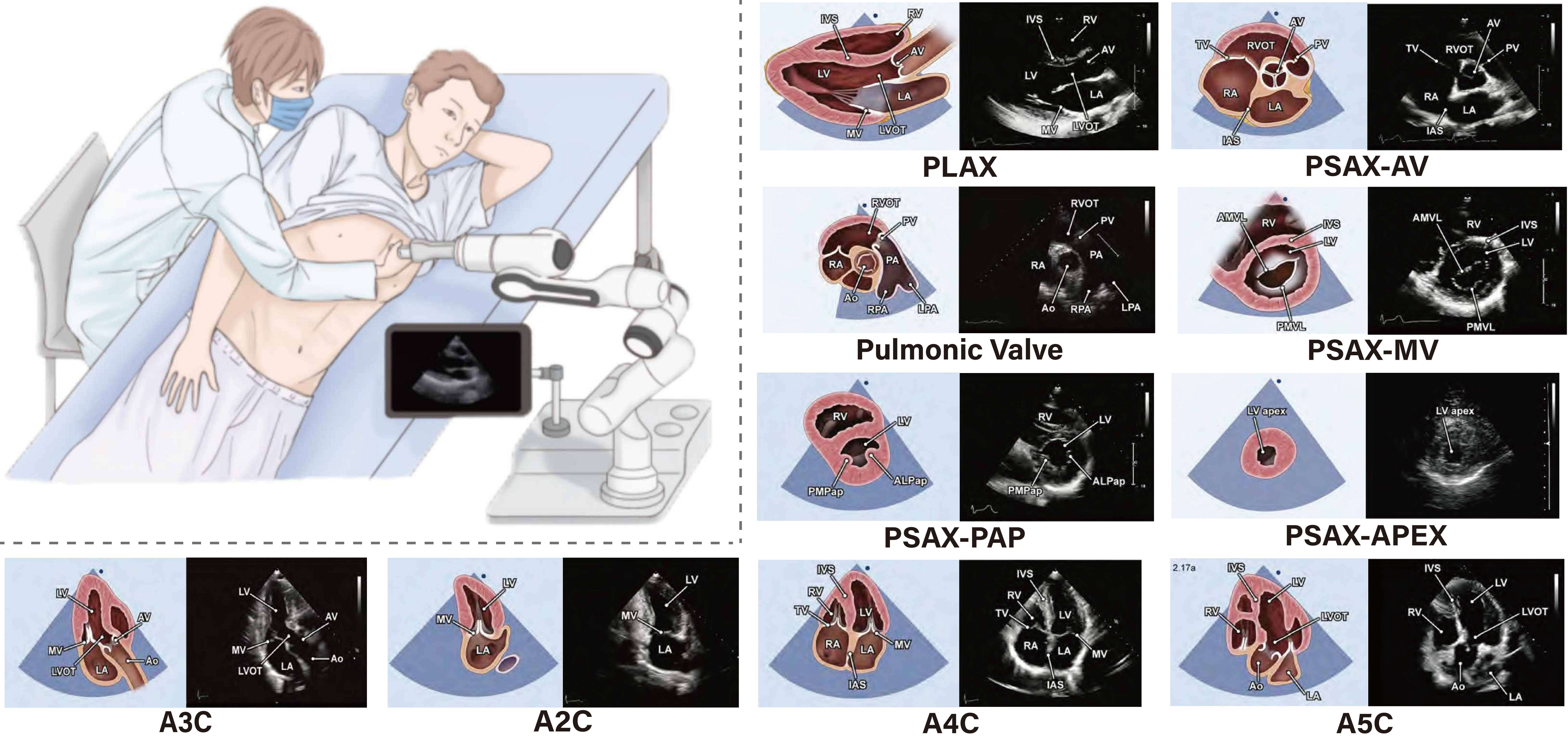}
\vspace{-18pt}
    \caption{Illustration of our dataset and task. Top-left: We collect expert demonstration data where the sonographer controls a robot arm with a probe, recording both image frames and probe motion synchronously. Remaining figure: The ten standard planes targeted for acquisition. Figures are adapted from \cite{mitchell2019guidelines,jiang2024cardiac}.}
    \vspace{-16pt}
    \label{fig:setting}
\end{figure}

During a typical echocardiographic examination, the sonographer sequentially captures several standard planes, measuring parameters in each view before forwarding the data for cardiologist interpretation. The probe guidance task, aimed at supporting or potentially automating this process, involves developing models that can direct the probe towards specified target planes. This automation could streamline the acquisition of accurate and diagnostically relevant images, making echocardiography more accessible and consistent across varied operator skill levels.


\begin{figure*}[ht]
    \centering
    \includegraphics[width=\linewidth]{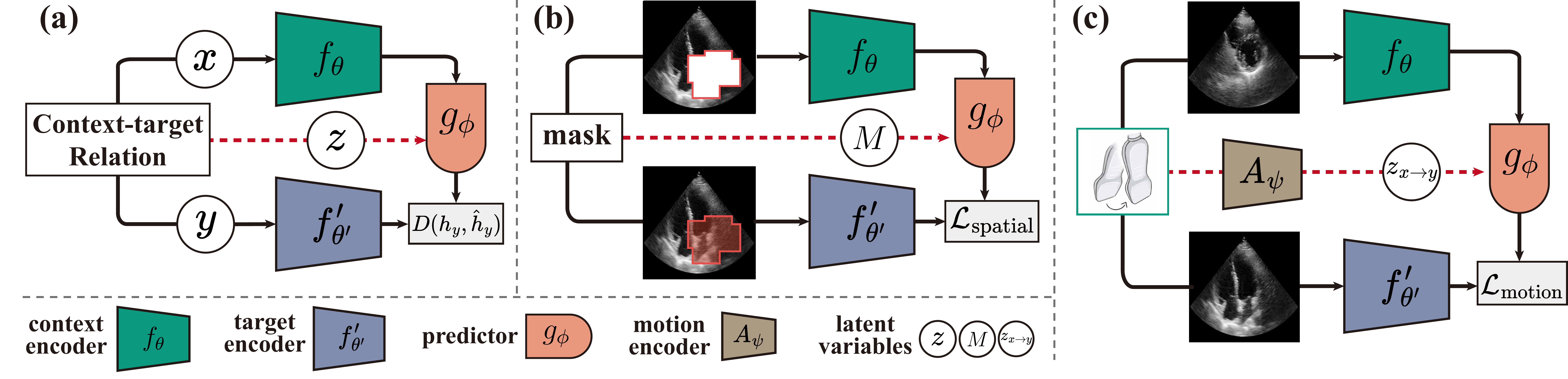}
    \vspace{-15pt}
    \caption{\textbf{Illustration of the world modeling tasks.} (a) A basic world modeling framework \cite{lecun2022path}, where the task is to predict the target $y$ from context $x$ in feature space, using a latent variable $z$ encoding their relationship. (b) The spatial world modeling task, which recovers masked anatomical structures. (c) The motion world modeling task, which predicts visual changes in the context based on probe motion.}
    \vspace{-15pt}
    \label{fig:method}
\end{figure*}

\subsection{Dataset and Task}
\label{sec:dataset}
Our study is conducted based on an expert demonstration dataset collected during routine clinical ultrasound exams, in which professional sonographers maneuvered an ultrasound probe mounted on a robotic arm. This setup enables synchronous recording of both image frames and probe pose information, as illustrated in Figure \ref{fig:setting}. A detailed introduction of the dataset can be found in the appendix.

The dataset is organized by \enquote{\textit{scans}}, each representing a recorded examination of a patient. A scan comprises a multi-minute echocardiography video (30 fps) and the probe’s corresponding pose for each frame in the anatomical coordinate system. Formally, each scan forms a visual-motion sequence $\{(\bm I_t, \bm p_t)\}_{t=1}^T$, where $T$ is the total number of timesteps, $\bm I_t$ is the ultrasound image at time $t$, and $\bm p_t$ denotes the probe pose. The probe poses are represented in six degrees of freedom (6-DOF): three translational coordinates (x, y, z) and three rotational components (yaw, pitch, roll) in Euler angles. Each probe pose belongs to the rigid transformation group, allowing us to express the \textit{relative movement}  between two poses $\bm p_i$ and $\bm p_j$ as $\bm p_{j\rightarrow i} = \bm p_i \cdot \bm p_j^{-1}$, where \enquote{$\cdot$} denotes the composition of transformations. While absolute probe poses can vary substantially across scans, the relative movements exhibit consistent patterns that reveal the sonographer's probe maneuvers.

During each scan, the sonographer visits multiple standard planes, with corresponding timestamps and plane types labeled as ground truth for the probe guidance task. Our goal is to predict the relative movement from each frame in the dataset to these target plane poses. Suppose the current timestep is $t_0$ and the target standard plane is reached at probe pose $\bm p^*$, then the ground truth movement at $t_0$ is given by $\bm a_{t_0}=\bm p^* \cdot \bm p_{t_0}^{-1}$. The probe guidance model leverages historical visual-motion data, $\{(\bm I_t, \bm p_t)\}_{t\le t_0}$, to predict the movement $\hat{\bm a}_{t_0}$ needed to reach the target plane. We focus on ten standard planes based on their clinical importance and prevalence in the dataset, as shown in Figure \ref{fig:setting}. 




\section{\methodname{}}
\label{sec:method}

This section presents our motion-aware world modeling framework for probe guidance, with the overall concept illustrated in Figure \ref{fig:idea}. We adopt a two-stage approach: a pre-training phase to construct a cardiac world model that captures essential cardiac ultrasound knowledge, followed by a fine-tuning phase with a motion-aware attention mechanism for integrating historical visual-motion data effectively.


\subsection{Pre-training Cardiac World Models}
Humans are believed to maintain internal models that capture complex patterns and dynamics of the world \cite{craik1967nature,ha2018world, lecun2022path}, encoding prior knowledge that supports perception, planning, and decision-making.
Similarly, an experienced sonographer develops a mental model of the heart’s structures and can anticipate visual changes as they adjust the probe. Inspired by this, we design world modeling tasks that equip our model with a similar understanding of cardiac anatomy and motion.


\textbf{A basic world modeling framework.} Our world model is based on the joint-embedding predictive architecture (JEPA) \cite{lecun2022path}, illustrated in Figure \ref{fig:method}(a). The model’s objective is to predict a target $y$, an unobserved portion of the world (such as the ultrasound scan in our study), using the available context $x$, conditioned on a latent variable $z$ that captures the relationship between the contexts and targets (\emph{e.g.} the probe movement that leads to the visual changes).

The JEPA consists of a \textit{context encoder} $f_\theta$ and a \textit{target encoder} $f'_{\theta'}$, which produce context and target features, $h_x=f_\theta(x)$ and $ h_y=f'_{\theta'}(y)$. A \textit{predictor} $g_\phi$ then predicts the target features conditioned on the latent variable $z$:
\begin{equation}
\setlength\abovedisplayskip{5pt}
\setlength\belowdisplayskip{5pt}
    \hat{h}(y) =  g_\phi(h_x, z) = g_\phi(f_\theta(x), z).
\end{equation}
The latent variable $z$ encodes essential information that bridges the context $x$ and target $y$, capturing the uncertainty and inherent dynamics of the world. 
The model’s objective is to minimize prediction error:
\begin{equation}
\setlength\abovedisplayskip{5pt}
\setlength\belowdisplayskip{5pt}
    \textnormal{minimize} \quad D(h_y, \hat{h}_y) = D(f'_{\theta'}(y), g_\phi(f_\theta(x);z)),
\end{equation}
where $D$ is an error function. To prevent feature collapse, the target encoder $f'_{\theta'}$ is an exponential moving average (EMA) of the context encoder. Below, we detail two world modeling tasks tailored for cardiac ultrasound.


\textbf{Spatial world modeling.}
Ultrasound images reveal a complex array of anatomical structures within the heart, including chambers, valves, and arteries, as shown in Figure \ref{fig:setting}. A thorough understanding of these structures’ layout and visual characteristics is essential for sonographers. To instill this anatomical knowledge in our model, we incorporate a mask-and-reconstruction task \cite{mae, ijepa, baevski2022data2vec, wang2023efficienttrain}, in which several contiguous regions of the image are masked, and the model is tasked with predicting the features for these areas. This task encourages the model to learn localized textures and the spatial relationships among anatomical structures.
Our model uses vision transformers as the context and target encoders, with masking achieved by dropping rectangular blocks of visual patches (Figure \ref{fig:method}(b)).
Only visible patches are passed to the context encoder, while the target encoder processes the entire image:
\begin{equation}
\setlength\abovedisplayskip{5pt}
\setlength\belowdisplayskip{5pt}
    h_x = f_\theta(\operatorname{Mask}(\bm I)), \quad h_y=f'_{\theta'}(\bm I).
\end{equation}
In the predictor, mask tokens $m$ carry the encoded positions of the masked regions, serving as the latent variable $z$. 
Letting $M$ denote the masked patch locations, the predictor output $\hat h_{y}$ is given by:
\begin{equation}
\setlength\abovedisplayskip{5pt}
\setlength\belowdisplayskip{5pt}
    \hat{h}_y = g_\phi\left(({h}_x + p_x)\oplus \left\{m+\operatorname{PE}(c)\ \right\}_{c\in M}\right),
\end{equation}
where $\oplus$ denotes concatenation, $p_x$ is the positional embedding of the context tokens, $c\in M$ are 2D patch coordinates, and $\operatorname{PE}(\cdot)$ represents sinusoidal positional encoding \cite{vaswani2017attention}. 
The loss is computed only over the masked locations:
\begin{equation}
\setlength\abovedisplayskip{5pt}
\setlength\belowdisplayskip{5pt}
\mathcal{L}_{\operatorname{spatial}} = \sum_{c\in M} \left\|g_\phi(f_\theta(x); M)_c - f'_{\theta'} (y)_c \right\|_1.
\end{equation}


\textbf{Motion world modeling.}
Beyond anatomical understanding in the 2D image plane, grasping the dynamics of ultrasound scanning is crucial for accurate probe guidance. As the probe is tilted, rotated, or moved, the visual features of cardiac structures—such as chamber alignment, valve orientation, and surrounding tissue—shift accordingly. Motion world modeling aims to capture the relationship between probe movements and these resulting visual changes. To build a motion-aware model, we introduce a predictive task in which the model anticipates changes in visual features based on probe movements. Specifically, given two images $\bm I_{a}, \bm I_{b}$ (serve as context and target) from the same scan, along with their relative probe movement $\bm p_{a\rightarrow b}$, we encode the motion information using a motion encoder $A_\psi$, yielding motion features $\bm z_{a\rightarrow b} = A_\psi(\bm p_{a\rightarrow b})$.  A single mask token carries these motion features, which are fed into the predictor along with the context features to predict the average-pooled target features:
\begin{equation}
\setlength\abovedisplayskip{5pt}
\setlength\belowdisplayskip{5pt}
\begin{aligned}
    \hat{h}_y &= g_\phi(f_\theta(\bm I_a); m+\bm z_{a\rightarrow b}),\\
    h_y &= \operatorname{AvgPool}(f'_{\theta'}(\bm I_b)).
\end{aligned}
\end{equation}
To enhance the model’s sensitivity to probe motion, we employ a contrastive objective \cite{sowrirajan2021moco}, where the prediction $\hat{h}_y$ and the corresponding target $h_y$ serve as positive pairs. Following \cite{chen2020simple}, both $\hat{h}_y,{h}_{y}$ are projected through a two-layer MLP\footnote{The projector is omitted in Eq \eqref{eq:nce} for brevity.} before applying the InfoNCE loss \cite {oord2018representation}:
\begin{equation}
\setlength\abovedisplayskip{5pt}
\setlength\belowdisplayskip{5pt}
\label{eq:nce}
    \mathcal L_{motion} = -\frac 1 B \sum_{i=1}^B\log \frac{\exp(\hat{h}_{y_i}^\top \cdot h_{y_i}/\tau)}{\sum_j \exp(\hat{h}_{y_i}^\top \cdot h_{y_j}/\tau)},
\end{equation}
where $B$ is the batch size.



\textbf{Joint Modeling.} \methodname{} integrates spatial and motion world modeling into a unified pre-training approach designed to maximize the complementary strengths of each task. Given two images $\bm I_a$ and $\bm I_b$ from the same scan, spatial world modeling is applied independently to both images, while motion world modeling uses $\bm p_{a\rightarrow b}$ to learn the transition between them. The final loss function is:
\begin{equation}
\setlength\abovedisplayskip{5pt}
\setlength\belowdisplayskip{5pt}
    \mathcal L_{\text{total}} = \mathcal L_{\text{spatial}} + \lambda \mathcal L_{\text{motion}},
\end{equation}
where $\lambda = 0.1$ balances the scale of the two losses.
Combining these two modeling strategies creates a holistic representation where each image is understood both in its anatomical context and as part of a dynamic process.


\subsection{Motion-Aware Probe Guidance}

\begin{figure}[t]
    \centering
    \includegraphics[width=0.8\columnwidth]{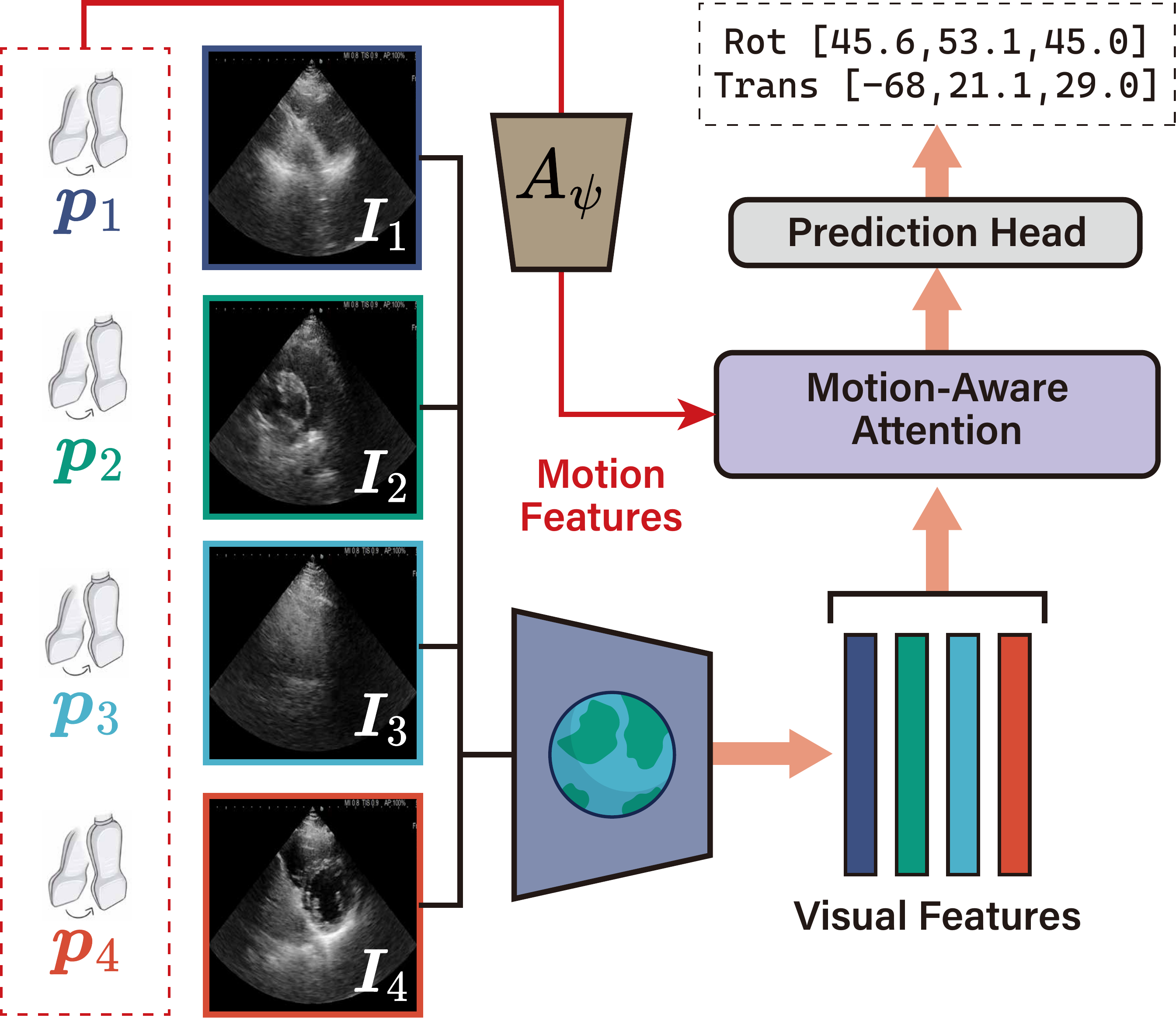}
    \vspace{-5pt}
    \caption{\textbf{The probe guidance pipeline.} Given a sequence of historical visual-motion pairs, we first extract features using the pre-trained visual and motion encoders. 
    These features are then integrated via a motion-aware attention mechanism and projected to the final guidance output.}
    \vspace{-15pt}
    \label{fig:pipeline}
\end{figure}



With the pre-trained world model in place, equipped with representations that capture both cardiac anatomy and visual-motion dynamics, we can now introduce a guidance predictor tailored for probe guidance applications. As shown in Figure \ref{fig:pipeline}, we introduce a motion-aware attention module that integrates historical visual-motion data by embedding motion information into attention features, enabling more effective aggregation of past observations.

Specifically, the probe guidance model predicts the probe movements to reach target planes by processing a sequence of visual-motion signal $\{I_{t_i}, p_{t_i}\}_{i=1}^N$, where $N$ is the sequence length, $t_1\le t_2\le\dots\le t_N$ with $t_N$ representing the latest timestep\footnote{Since the original videos contain thousands of images, we sample a subset of timesteps as input to the model.}. Previous methods typically organize historical data into an interleaved image-action sequence $\{\bm I_{t_1}, \bm p_{t_1\rightarrow t_2}, \bm I_{t_2},\bm p_{t_2\rightarrow t_3},\bm I_{t_3},\dots,\bm I_{t_N}\}$ and encode the sequence using a recurrent model. However, we argue that this formulation is suboptimal as it fails to fully utilize the rich motion data available. Instead, we propose a motion-aware attention mechanism that incorporates motion signals directly into the interactions between all frame tokens. This approach aligns the attention mechanism with the 3D spatial transformations of the probe, facilitating more precise modeling of the spatial relationships between frames.

\begin{figure}[t]
    \centering
    \includegraphics[width=\columnwidth]{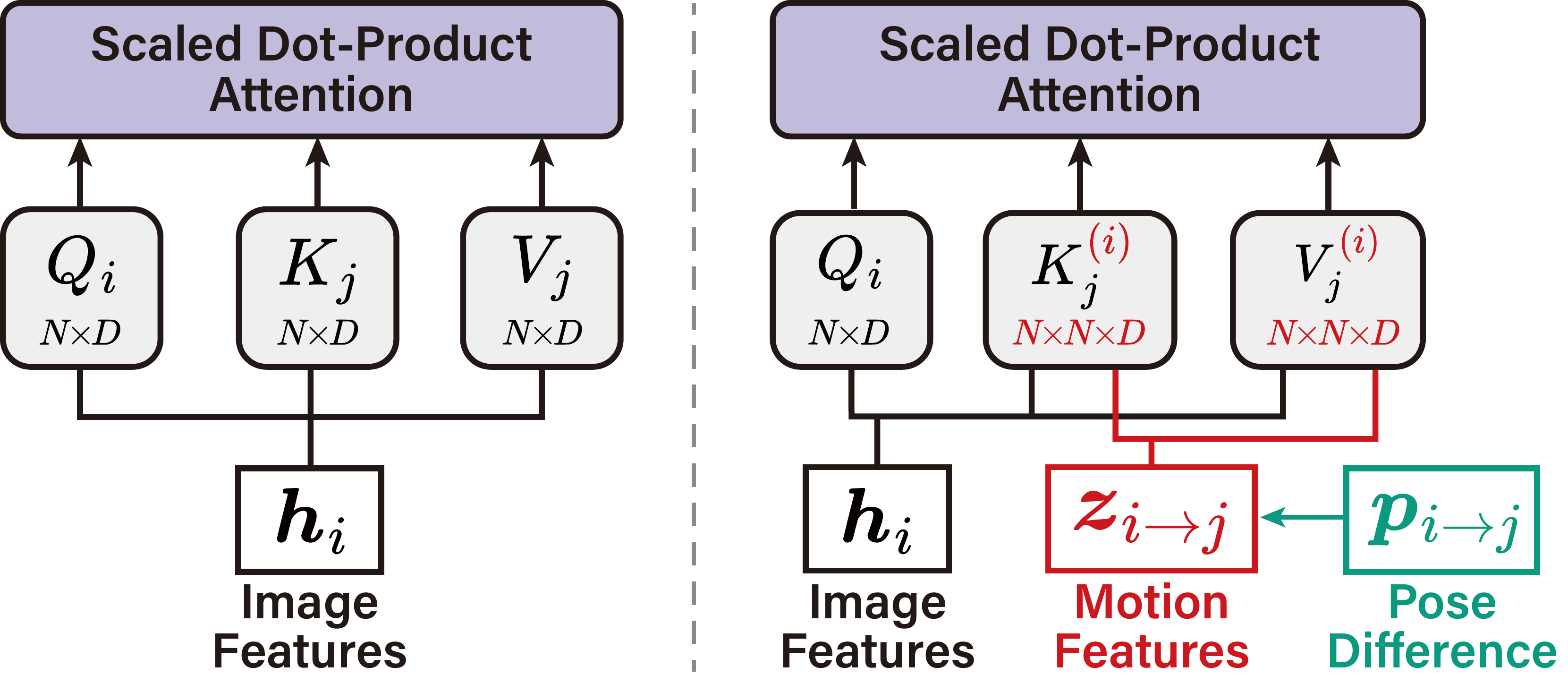}
    \vspace{-17pt}
    \caption{\textbf{Comparison of standard (left) and motion-aware (right) attention,} where the latter incorporates pairwise pose differences across the tokens into the key-value pairs.}
    \vspace{-20pt}
    \label{fig:attn}
\end{figure}

Specifically, we first use the pre-trained world model to extract the features of the visual and motion data:
\begin{equation}
\setlength\abovedisplayskip{5pt}
\setlength\belowdisplayskip{5pt}
    \bm h_i = \operatorname{AvgPool}(f_\theta(\bm I_{t_i})),\quad \bm z_{i\rightarrow j} = A_\psi(\bm p_{{t_i}\rightarrow {t_j}}).
\end{equation}
The average-pooled visual features $\{\bm h_i\}_{i=1}^N$ serve as input tokens to our motion-aware attention module. The standard scaled dot-product attention is computed as follows:
\begin{equation}
\setlength\abovedisplayskip{5pt}
\setlength\belowdisplayskip{5pt}
    \label{eq:attn}
    O_i = \sum_{j} \frac{\exp(Q_i^\top K_j)}{\sum_l \exp(Q_i^\top K_l)} V_j,
\end{equation}
where $O_i$ is the output representation of token $i$. Since this operation is permutation invariant, position information must be embedded in the tokens for the model to understand spatial relationships. In general transformer architectures, this is achieved with absolute \cite{vaswani2017attention} or relative \cite{shaw2018self,su2024roformer} positional embeddings. 
For probe guidance, however, it is essential to embed motion-specific information into the attention mechanism to capture the relative spatial transformations of the probe. We achieve this by encoding the pairwise pose differences into the key and value features:
\begin{equation}
\setlength\abovedisplayskip{5pt}
\setlength\belowdisplayskip{5pt}
    K_j^{\textcolor{BrickRed}{(i)}} = \operatorname{MLP}(\bm h_j, \bm z_{i\rightarrow j}),\ V_j^{\textcolor{BrickRed}{(i)}} = \operatorname{MLP}(\bm h_j, \bm z_{i\rightarrow j}).
\end{equation}
This allows each query token to be associated with a unique set of key-value pairs that encode the relative pose information. The attention output for token $i$ is computed as:
\begin{equation}
\setlength\abovedisplayskip{5pt}
\setlength\belowdisplayskip{5pt}
     \label{eq:attn_motion}
    O_i = \sum_{j} \frac{\exp(Q_i^\top K_j^{\textcolor{BrickRed}{(i)}})}{\sum_l \exp(Q_i^\top K_l^{\textcolor{BrickRed}{(i)}})} V_j^{\textcolor{BrickRed}{(i)}}.
\end{equation} 
This motion-aware approach contrasts with standard self-attention by incorporating relative motion information, enabling better capture of spatial changes induced by probe movements. Additionally, we employ a multi-head design \cite{vaswani2017attention} to capture diverse attention patterns.
Figure \ref{fig:attn} compares the standard and motion-aware attention mechanisms.

After the motion-aware attention, the aggregated features are fed into a prediction head to regress the probe movement to the target plane. The guidance loss is computed as $\mathcal L_{\text{guide}} = \|\bm a_{t} - \hat{\bm a}_{t}\|_1$,
where $\hat{\bm a}_{t}$ is the model’s predicted movement to the target plane at timestep $t$ and $\bm a_{t}$ is the ground truth movement.

\newcolumntype{P}[1]{>{\centering\arraybackslash}p{#1}}

\begin{figure*}[ht]
    \centering
    \includegraphics[width=0.8\linewidth]{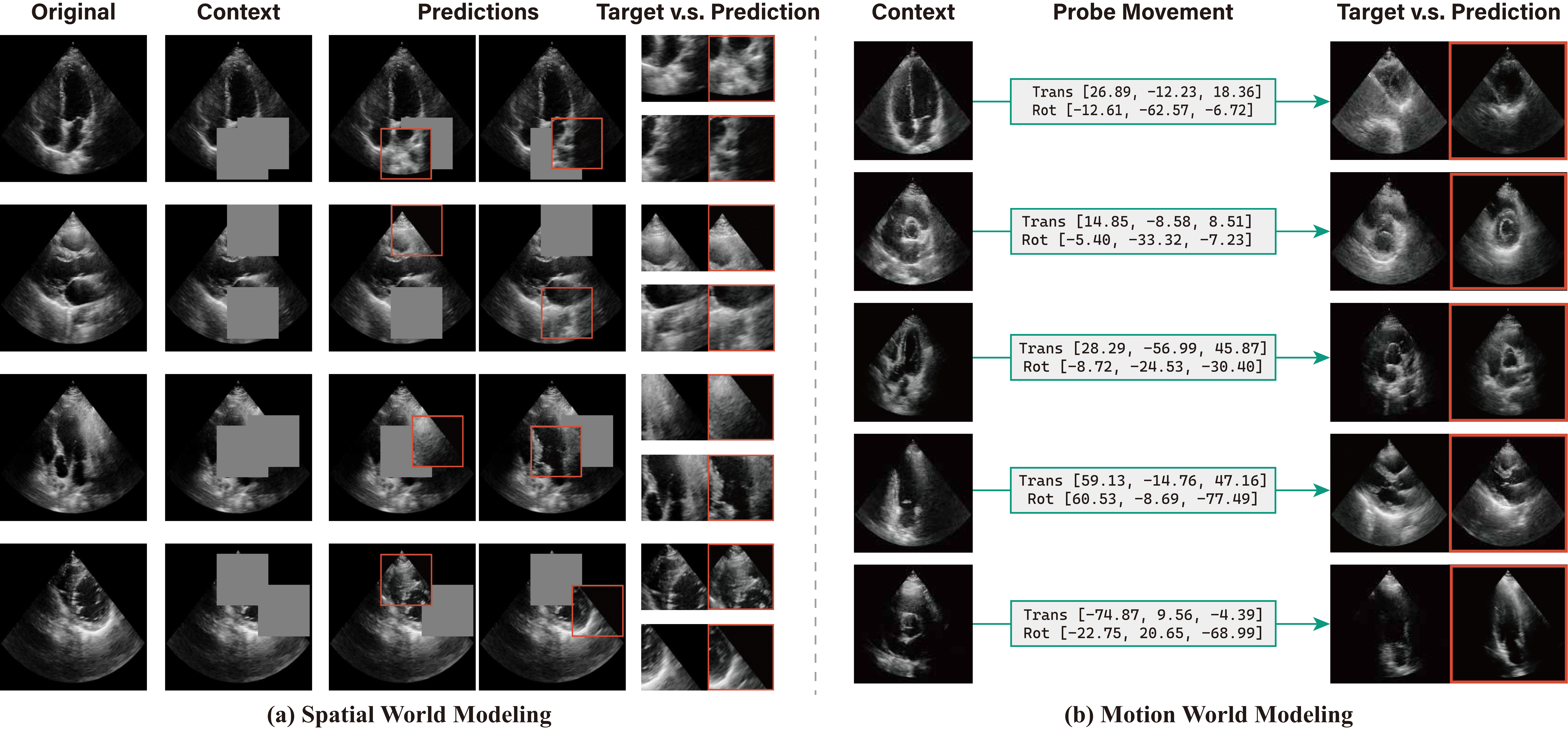}
    \vspace{-10pt}
    \caption{\textbf{Visualization of world modeling predictions.}  Using a diffusion model, we map the outputs of the predictor to pixel space. The predictor effectively recovers (a) masked anatomical regions and (b) the visual changes resulting from probe movement. The predicted regions are highlighted with red bounding boxes.}
    \vspace{-5pt}
    \label{fig:vis_world}
\end{figure*}

\begin{table*}[ht]
    \centering
    \caption{\textbf{Performance comparison of \methodname{} and baselines on the probe guidance task}. Experiments are conducted under both the single-frame and sequential protocols.  The mean absolute error is reported. \enquote{Trans.} denotes the translation error (x, y, z) in millimeters, while \enquote{Rot.} represents the rotation error (yaw, pitch, roll) in degrees. The best two results are \textbf{bold-faced} and \underline{underlined}, respectively. 
    }
    \label{tab:result_main}
    \vspace{-10pt}
    \resizebox{1.0\linewidth}{!}{
    \centering
    \begin{tabular}{p{2.7cm}*{20}{P{0.5cm}}P{1cm}}
    \toprule
        \multirow{2}{*}{Method} & \multicolumn{2}{c}{\footnotesize PLAX} & \multicolumn{2}{c}{\footnotesize PSAX-AV} & \multicolumn{2}{c}{\footnotesize PSAX-PV} & \multicolumn{2}{c}{\footnotesize PSAX-MV} & \multicolumn{2}{c}{\footnotesize PSAX-PAP} & \multicolumn{2}{c}{\footnotesize PSAX-APEX}  & \multicolumn{2}{c}{\footnotesize A4C} & \multicolumn{2}{c}{\footnotesize A5C} & \multicolumn{2}{c}{\footnotesize A3C} & \multicolumn{2}{c}{\footnotesize A2C} &\multirow{2}{*}{\textcolor{BrickRed}{Avg}}\\
        & {\footnotesize Trans.} & {\footnotesize Rot.} & {\footnotesize Trans.} & {\footnotesize Rot.}& {\footnotesize Trans.} & {\footnotesize Rot.}& {\footnotesize Trans.} & {\footnotesize Rot.}& {\footnotesize Trans.} & {\footnotesize Rot.}& {\footnotesize Trans.} & {\footnotesize Rot.}& {\footnotesize Trans.} & {\footnotesize Rot.}& {\footnotesize Trans.} & {\footnotesize Rot.}& {\footnotesize Trans.} & {\footnotesize Rot.}& {\footnotesize Trans.} & {\footnotesize Rot.}  & \\
        \midrule
        \multicolumn{21}{l}{\textit{Single-frame protocol, not pre-trained on ultrasound data}} &  \\
        Scratch \cite{vit} & 8.84 &  7.78 & 8.12 & 8.31 & 8.65 & 9.02 & 8.18 & 9.21 & 8.05 & 9.13 & 9.56 & 9.89 & 8.86 & 7.95 & 9.06 & 10.04 & 9.35 & 10.22 & 9.09 & 11.97 & \textcolor{BrickRed}{9.07} \\
        DeiT \cite{touvron2021training}& 8.50 & 7.27 & 8.09 & 8.02 & 8.51 & 8.71 & 7.90 & 8.66 & 7.85 & 8.62 & 9.21 & 9.23 & 8.43 & 7.37 & 8.55 & 9.54 & 8.70 & 9.71 & 8.52 & 11.28 & \textcolor{BrickRed}{8.63} \\
        DINOv2 \cite{oquab2023dinov2}& 8.31 & 7.21 & 7.82 & 7.82 & 8.36 & 8.75 & 7.70 & 8.66 & 7.66 & 8.42 & 9.09 & 9.04 & 8.40 & 7.27 & 8.52 & 9.41 & 8.70 & 9.38 & 8.62 & 11.27 & \textcolor{BrickRed}{8.52} \\
        \midrule
        \multicolumn{21}{l}{\textit{Single-frame protocol, pre-trained on ultrasound data}} & \\
        BioMedCLIP \cite{zhang2023biomedclip}& 8.40 & 7.33 & 7.89 & 8.00 & 8.44 & 8.84 & 7.79 & 8.88 & 7.78 & 8.85 & 9.16 & 9.38 & 8.62 & 7.81 & 8.73 & 10.04 & 8.88 & 9.59 & 8.75 & 11.62 & \textcolor{BrickRed}{8.74}   \\
        LVM-Med \cite{mh2024lvm}& 8.55 & 7.26 & 8.00 & 7.94 & 8.42 & 8.62 & 7.95 & 8.81 & 7.89 & 8.85 & 9.31 & 9.27 & 8.70 & 7.59 & 8.88 & 9.65 & 9.07 & 9.65 & 8.80 & 11.34 & \textcolor{BrickRed}{8.73} \\
        US-MoCo \cite{mocov3} & 8.75 & 7.46 & 8.08 & 7.90 & 8.29 & 8.64 & 7.95 & 8.76 & 7.90 & 8.64 & 9.22 & 9.23 & 8.55 & 7.40 & 8.78 & 9.76 & 8.88 & 9.70 & 8.74 & 11.55 & \textcolor{BrickRed}{8.71} \\
        US-MAE \cite{mae} & 8.31 & 7.11 & \underline{7.74} & 7.61 & 8.30 & \underline{8.23} & 7.69 & 8.55 & 7.70 & 8.56 & \textbf{9.02} & 9.05 & 8.35 & 7.26 & 8.47 & 9.67 & 8.55 & 9.46 & 8.45 & 11.05 & \textcolor{BrickRed}{8.46} \\
        USFM \cite{jiao2024usfm}& 8.34 & 7.14 & \underline{7.74} & 7.69 & \underline{8.28} & 8.51 & 7.62 & \underline{8.38} & 7.69 & \underline{8.41} & \textbf{9.02} & \underline{9.00} & 8.20 & 7.18 & 8.33 & \underline{9.38} & \underline{8.51} & \underline{9.37} & 8.39 & 11.14 & \textcolor{BrickRed}{8.42}  \\
        EchoCLIP \cite{christensen2024vision}& \underline{8.29} & \underline{6.86} & 7.78 & \underline{7.52} & 8.45 & 8.51 & \underline{7.53} & 8.44 & \underline{7.60} & 8.47 & 9.17 & 9.18 & \underline{8.16} & \underline{7.03} & \underline{8.22} & \textbf{9.31} & 8.58 & \textbf{8.88} & \underline{8.35} & \textbf{10.97} & \textcolor{BrickRed}{\underline{8.37}} \\
        \methodname{}& \textbf{7.95} & \textbf{6.83} & \textbf{7.45} & \textbf{7.37} & \textbf{7.86} & \textbf{7.99} & \textbf{7.37} & \textbf{7.97} & \textbf{7.29} & \textbf{7.85} & \underline{9.04} & \textbf{8.57} & \textbf{7.99} & \textbf{7.00} & \textbf{8.09} & 9.42 & \textbf{8.27} & 9.59 & \textbf{8.22} & \underline{10.99} & \textcolor{BrickRed}{\textbf{8.15}}  \\
        \midrule
        \midrule
        \multicolumn{21}{l}{\textit{Sequential protocol, using the same visual backbone}} & \\
        
       US-GuideNet \cite{droste2020automatic}& 7.14 & 6.13 & 7.61 & 7.35 & 8.42 & 7.91 & 6.56 & 7.62 & 7.18 & 7.77 & 8.34 & 8.31 & 7.66 & 6.31 & 7.55 & 7.14 & 8.32 & 8.16 & 8.10 & 10.87 & \textcolor{BrickRed}{7.72} \\
        Decision-T \cite{chen2021decision} & \underline{6.72} & \underline{5.78} & \underline{7.11} & \underline{6.98} & \underline{8.12} & \textbf{7.46} & 6.46 & \underline{7.47} & \underline{6.53} & \textbf{7.25} & 8.77 & \underline{8.09} & \underline{7.07} & 6.02 & 
        \underline{6.79} & \underline{6.87} & 7.83 & 8.48 & 7.78 & 11.22 & \textcolor{BrickRed}{7.44} \\
        Seq-aware \cite{jiang2024sequence}& 6.99 & 5.87 & 7.42 & 7.06 & 8.38 & 7.89 & \underline{6.11} & 7.74 & 6.68 & 7.61 & \textbf{7.63} & 8.19 & 7.19 & \underline{5.89} & 6.98 & 6.88 & \underline{7.74} & \underline{7.80} & \underline{7.70} & \underline{10.59} & \textcolor{BrickRed}{\underline{7.42}} \\
        \methodname{}& \textbf{6.28} & \textbf{5.40} & \textbf{6.92} & \textbf{6.94} & \textbf{7.95} & \underline{7.47} & \textbf{5.78} & \textbf{7.21} & \textbf{6.28} & \underline{7.38} & \underline{7.71} & \textbf{7.71} & \textbf{6.74} & \textbf{5.58} & \textbf{6.46} & \textbf{6.64} & \textbf{7.36} & \textbf{7.50} & \textbf{7.23} & \textbf{10.47} & \textcolor{BrickRed}{\textbf{7.05}} \\ 
        \bottomrule
    \end{tabular}}
    \vspace{-10pt}
\end{table*}

\section{Experiments}
In this section, we first demonstrate \methodname{}'s world modeling capabilities via visualizations. We then compare \methodname{} with existing methods on the probe guidance task using two evaluation protocols: single-frame and sequential. Finally, ablation studies highlight the contribution of each framework component.

\textbf{Implementation details.} Our dataset comprises 356 routine cardiac ultrasound scans, collected and processed as described in Section \ref{sec:dataset}. The dataset consists of approximately one million image frames, each paired with its corresponding probe pose. The data is divided into 284 scans for training and 72 scans for testing, with no overlap in patients between the two splits. 
For pre-training, we employ a ViT-Small \cite{vaswani2017attention} as the visual backbone and use a transformer with six blocks as the predictor. 
Target encoder $f'_{\theta'}$ is the exponential moving average of the context encoder $f_\theta$. 
Training is conducted with a batch size of 1024 over 300 epochs on four A100 GPUs.
After pre-training, the model is fine-tuned using ground truth movements toward the standard planes.

\subsection{\methodname{} as a World Model}


To validate the world modeling capabilities of \methodname{}, we train a generative model that maps the predicted features $\hat{h}_y$ back to pixel space using the RCDM framework \cite{bordes2021high}. The context-target pairs are constructed in the same manner as in the spatial and motion world modeling tasks. A diffusion model is then trained to generate the target image conditioned on the predicted features $\hat{h}_y$ (average pooled for spatial modeling).
As shown in Figure \ref{fig:vis_world}(a), \methodname{} successfully recovers masked regions that are consistent with the context, demonstrating its understanding of the appearance and spatial relationship between anatomical structures.
In Figure \ref{fig:vis_world}(b), \methodname{} simulates ultrasound scanning by predicting visual changes based on probe movements. This ability highlights the model’s capacity to capture the interaction between visual and motion signals, which is crucial for guiding the probe toward the destination.


\subsection{\methodname{} for Probe Guidance}
\label{sec:result}

    

\begin{figure*}[t]
    \begin{center}
    \begin{minipage}{0.7\columnwidth}
        \centering
    \includegraphics[width=1.0\columnwidth]{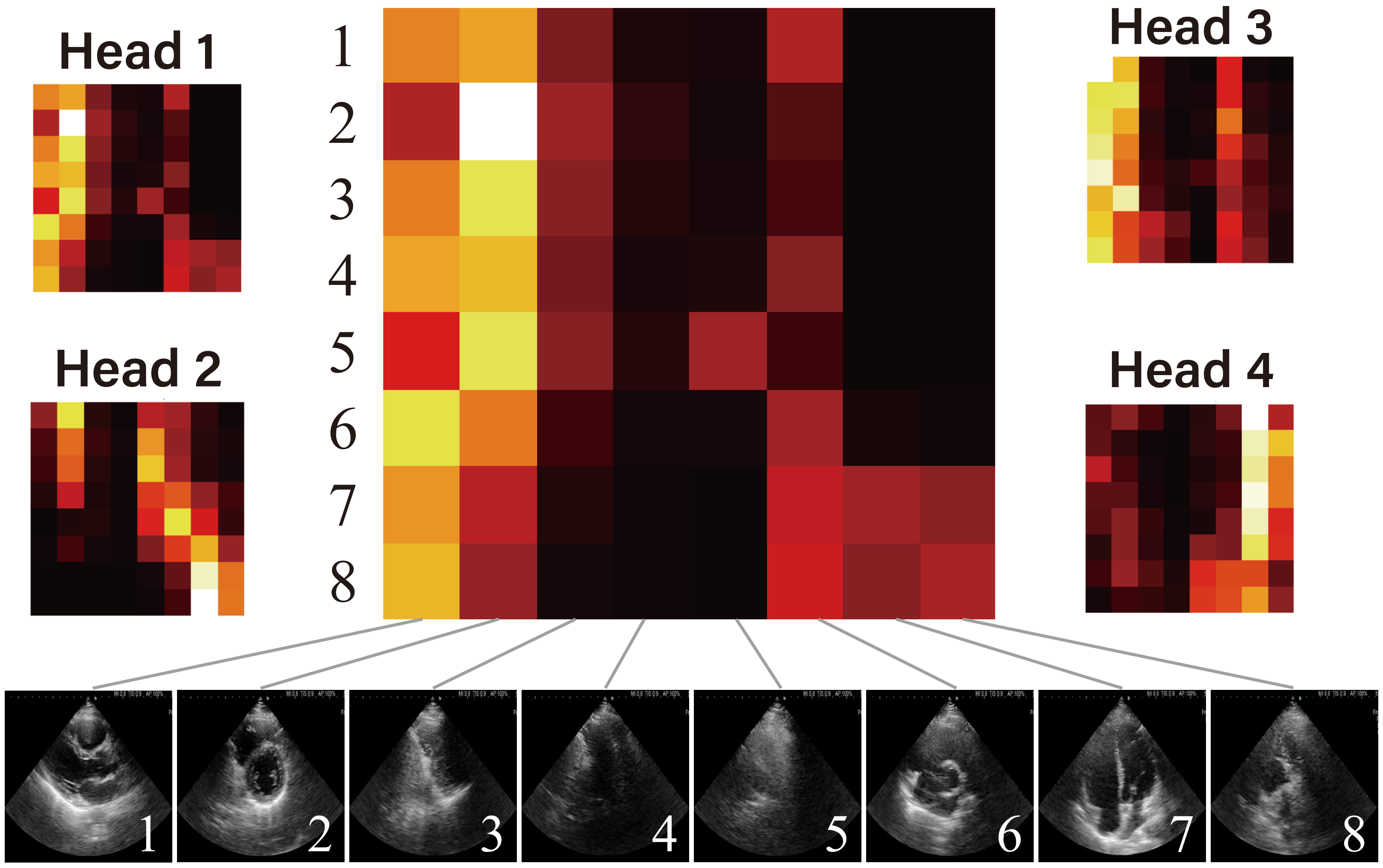}
    \vspace{-15pt}
    \caption{\textbf{Visualization of attention scores.}}
    \label{fig:vis_attn}
    \end{minipage}
    \hspace{0.05in}
    \begin{minipage}{0.65\columnwidth}
    \centering
    \captionof{table}{\textbf{Effectiveness of the world modeling tasks.} This evaluation uses the single-frame protocol. \enquote{Trans.} and \enquote{Rot.} refer to the translation and rotation error averaged over the ten standard planes.}
    \label{tab:single_abl}
    \vspace{-5pt}
    \resizebox{1.0\linewidth}{!}{
    \begin{tabular}{cc|ccc}
    \toprule
        Spatial &  Motion & Trans. & Rot. & Avg\\
        \midrule
      \ding{55} & \ding{55} & 8.77 & 9.37 & 9.07 \\
      \ding{55} & \ding{51} & 8.39 & 8.84 & 8.62 \\
     \ding{51} &  \ding{55} & 8.16 & 8.60 & 8.38 \\
       \ding{51} & \ding{51} & \textbf{7.95} & \textbf{8.36} & \textbf{8.15} \\
        \bottomrule
    \end{tabular}
    }
    \end{minipage}
    \hspace{0.05in}
    \begin{minipage}{0.65\columnwidth}
    \centering
    \captionof{table}{\textbf{Ablation study on motion-awareness.} This analysis uses the sequential protocol. \enquote{Motion} indicates whether pose information is available.}
    \label{tab:seq_abl}
    \vspace{5pt}
    \resizebox{1.0\linewidth}{!}{
    \begin{tabular}{c|c|ccc}
    \toprule
        Backbone & Motion & Trans. & Rot. & Avg\\
        \midrule
        \multirow{2}{*}{DeiT \cite{touvron2021training}}& \ding{55} &  8.48 & 8.76 & 8.62 \\
        & \ding{51} &  7.42 & 7.65 & 7.53 \\
        \midrule
       \multirow{2}{*}{\methodname{}} &\ding{55} & 7.91 & 8.06 & 7.98 \\
       & \ding{51} & \textbf{6.87} & \textbf{7.23} & \textbf{7.05} \\
        \bottomrule
    \end{tabular}}
        \end{minipage}
    \end{center}
    \vspace{-25pt}
 \end{figure*}

\textbf{Evaluation protocol.} Evaluating different approaches in real-person experiments is both costly and time-consuming. To address this, we propose two evaluation protocols based on the collected scanning data, as detailed below:

\emph{Single-frame protocol}: In this protocol, a two-layer MLP head is attached to each visual backbone to predict probe movements to ten target planes from a single ultrasound image. This task focuses on evaluating the representation power of the visual backbones, ensuring that all models are compared using the same prediction head and receive identical supervision signals. We report the mean absolute prediction error averaged across all frames in the test set.

\emph{Sequential protocol}:  This protocol simulates an online deployment scenario, where the model predicts probe movements toward \emph{unvisited} planes based on historical visual-motion data\footnote{To manage the long frame sequence, we sample $N$ timesteps from the history following a decayed density rate. This allows the model to retain long-term context without losing focus on the current timestep.}. It provides a more comprehensive evaluation of the probe guidance framework's performance.
Specifically, for a scan at timestep $t$, the model uses $N$ historical visual-motion pairs $\{\bm I_{t_i},\bm p_{t_i}\}_{i=1}^N$, where $t_i\le t, t_N=t$. In this setup, the model predicts movements toward planes that have not been visited by timestep $t$ 
. To ensure a symmetric evaluation, we evaluate the model in both the forward and reversed directions for each scan. The mean absolute prediction error is reported, averaged over every timestep across 30 test scans where all planes are visited. To ensure a fair comparison, the \emph{same} visual backbone (pre-trained using our proposed method) is used for all baselines and our method.




\textbf{Baselines.} In the single-frame protocol, we compare \methodname{} against a diverse set of pre-trained models, including general domain vision models such as DeiT \cite{touvron2021training} and DINOv2 \cite{oquab2023dinov2}, as well as medical vision models like BioMedCLIP \cite{zhang2023biomedclip} and LVM-Med \cite{mh2024lvm}, ultrasound-specific models USFM \cite{jiao2024usfm} and EchoCLIP \cite{christensen2024vision}, as well as two representative self-supervised learning methods applied to our ultrasound dataset, namely US-MoCo and US-MAE, derived from MoCo \cite{mocov3} and MAE \cite{mae}. In the sequential protocol, we evaluate \methodname{} against existing probe guidance frameworks, including US-GuideNet \cite{droste2020automatic} and Sequence-aware \cite{jiang2024sequence}, as well as a sequential decision-making model, Decision Transformer \cite{chen2021decision}. These baseline frameworks are reproduced using our own dataset to ensure a fair comparison. 
Additional details on the implementation of these baselines are provided in the appendix.

\textbf{Results on the single-frame protocol} are shown in the upper section of Table \ref{tab:result_main}. Our method consistently outperforms all baselines in predicting the probe movement for acquiring the ten standard planes from individual images. \methodname{} achieves the lowest error in 16 out of 20 translation/rotation errors across each subtask, outperforming the best alternative, EchoCLIP, by 0.22 in the averaged mean absolute error. Ultrasound-specific pre-trained models, such as USFM, EchoCLIP, and US-MAE, generally outperform general-purpose models, emphasizing the importance of in-domain transfer.

\textbf{Results on sequential protocol} are presented in the lower section of Table \ref{tab:result_main}. \methodname{} surpasses existing probe guidance frameworks while using the same visual backbone initialization across all models.
This suggests that the novel guidance prediction module, which incorporates motion-aware attention, enables more effective use of visual-motion data than previous paradigms.

\subsection{Analytical Results}
\textbf{Visualization of attention patterns.} 
To better understand the motion-aware attention mechanism, we visualize the attention scores between a set of 8 ultrasound frames along with their associated motion information. As shown in Figure \ref{fig:vis_attn}, different attention heads capture diverse attention patterns, where the model tends to focus on high-quality frames (e.g., frames 1 and 2) while ignoring less informative ones (e.g., frames 3, 4, and 5). This demonstrates the model’s ability to selectively aggregate useful information for improved probe guidance.


\begin{figure}[h]
    \centering
    \includegraphics[width=0.85\linewidth]{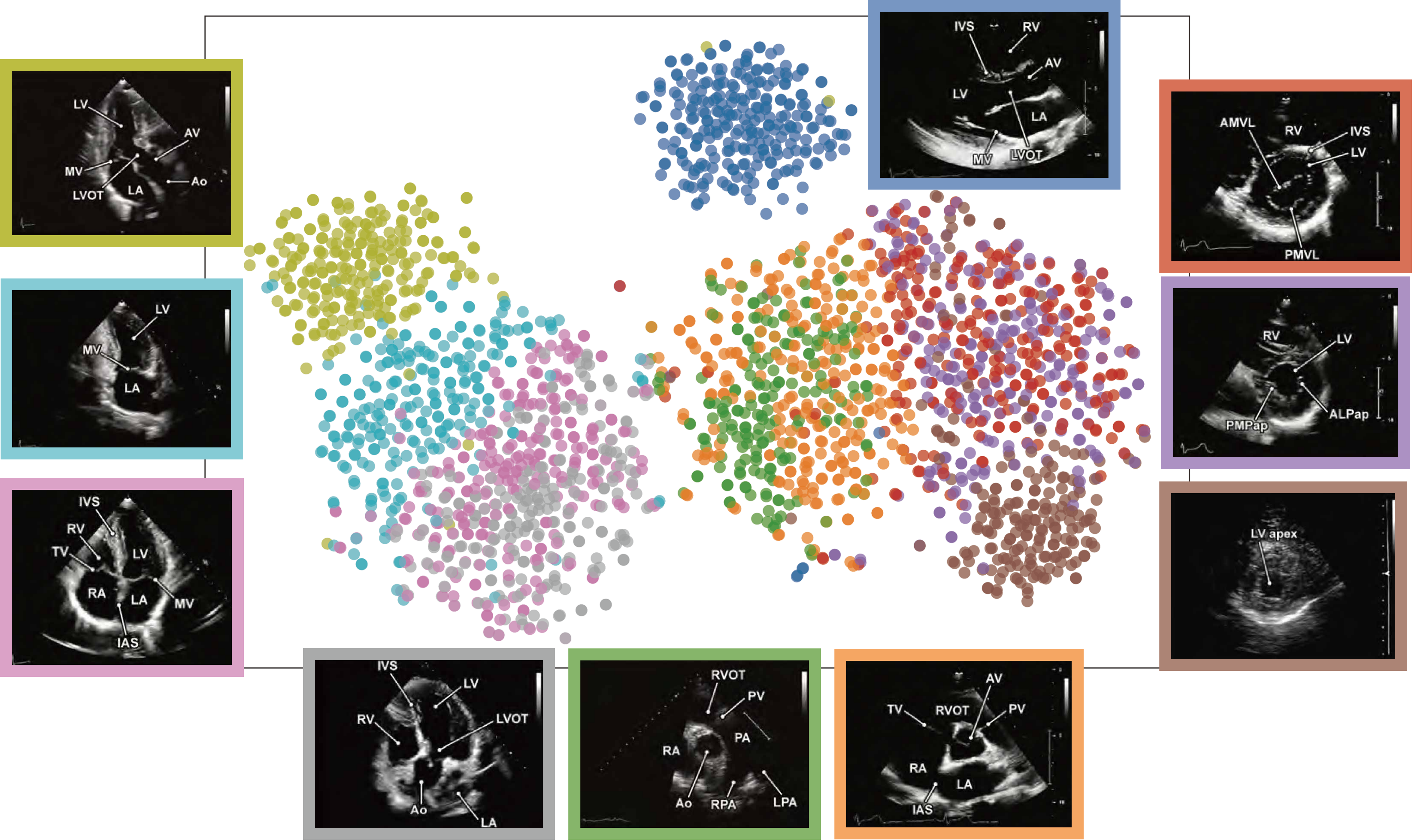}
    \vskip -0.08in
    \caption{\label{fig:tsne}\textbf{t-SNE visualization.}}
    \vspace{-20pt}
\end{figure}

\textbf{Visualization of the plane features.} We visualize the image features of the ten standard planes in the dataset using t-SNE. As shown in Figure \ref{fig:tsne}, images representing the same type of plane cluster together, with similar planes appearing closer to each other, indicating that the model learns meaningful semantics relevant to the downstream task.


\textbf{Ablation study of the world modeling tasks} is shown in Table \ref{tab:single_abl}. \methodname{} benefits from combining both spatial and motion modeling, leading to a significant improvement in representation quality.


\textbf{Ablation study on motion-awareness} is shown in Table \ref{tab:seq_abl}. Our results show that leveraging motion signals significantly reduces guidance errors. Additionally, when using the pre-trained backbone, the motion-aware guidance module provides a further performance boost, demonstrating the coherence and effectiveness of the proposed framework.

\section{Conclusion}

This paper proposed \methodname{}, a motion-aware world modeling framework designed for probe guidance in echocardiography. 
By integrating spatial and motion-aware world modeling tasks, our pre-trained encoder captures rich representations of ultrasound knowledge, encompassing anatomical structures as well as motion-driven dynamic visual changes. 
A motion-aware attention mechanism further enhances probe guidance by seamlessly incorporating the visual-motion features.
Our model outperforms existing pre-trained models and frameworks in predicting probe movements for acquiring standard planes. 
Our work provides a principled approach to modeling medical images, which considers both the medical knowledge and the inherent dynamics of imaging processes, advancing machine learning models for embodied medical applications.



\section*{Acknowledgements} 
The work is supported in part by the National Key R\&D Program of China under Grant 2024YFB4708200.

{
    \small
    \bibliographystyle{ieeenat_fullname}
    \bibliography{main}
}


\clearpage
\setcounter{page}{1}
\maketitlesupplementary
\renewcommand{\thefigure}{A\arabic{figure}}

\setcounter{figure}{0}
\appendix

\section{Dataset}
The echocardiography dataset used in this study was collected during routine clinical examinations, where certified sonographers performed ultrasound scans (M5S probe, GE Vivid E7 machine) using a probe mounted on a Franka Panda robot arm. During each scan, both the ultrasound videos (30 fps) and the corresponding probe poses were simultaneously recorded. All subjects in the dataset were healthy adult males. The data collection process was conducted in compliance with ethical guidelines and was reviewed and approved by the relevant institutional ethics committee.

This study utilizes a subset of 356 scans curated from the dataset, comprising approximately one million images and corresponding ultrasound probe poses. 
Each scan lasts several minutes, during which the sonographer maneuvers the probe and examines the heart from various views.

\begin{figure}[h]
    \centering
    \includegraphics[width=0.9\linewidth]{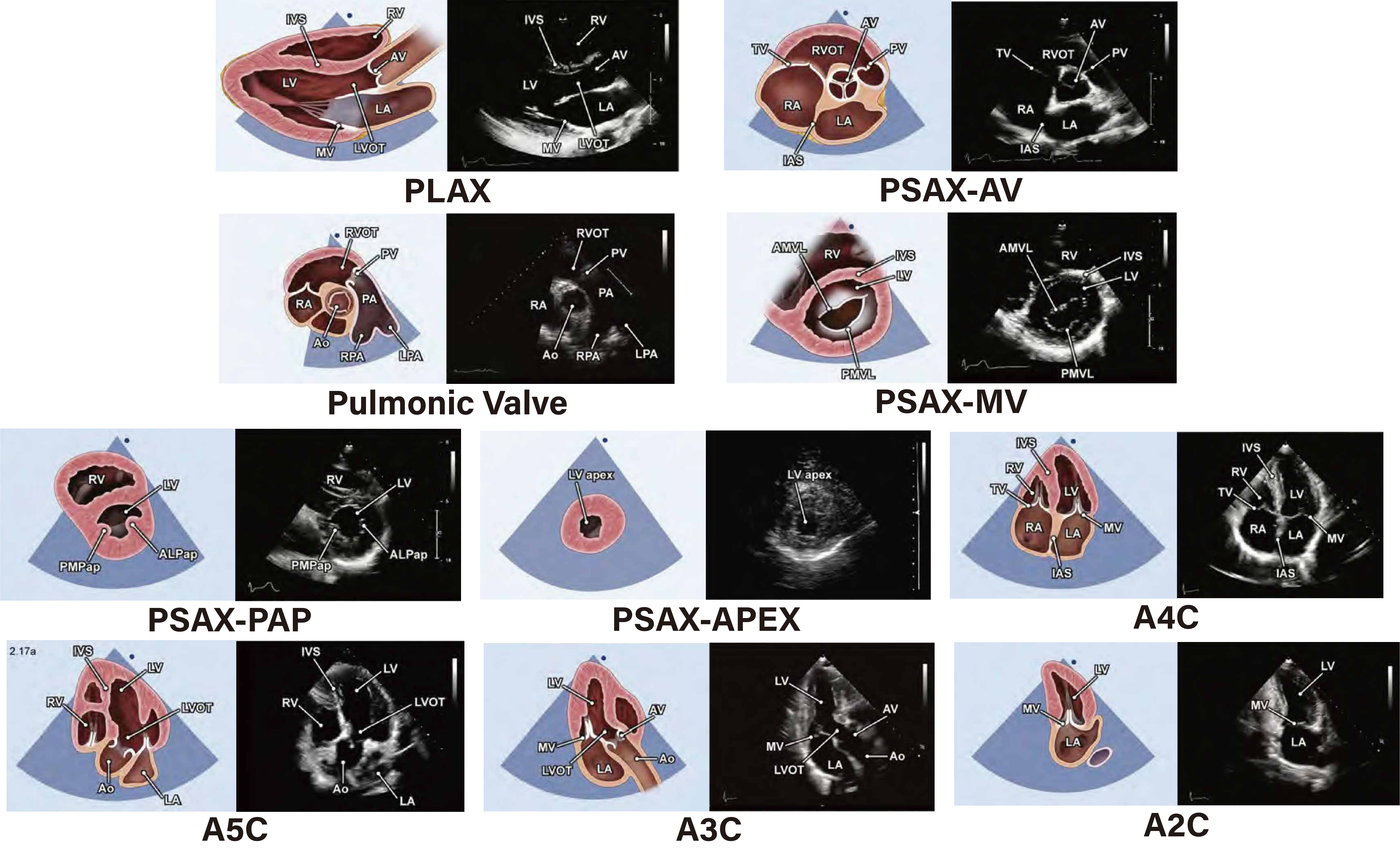}
    \vskip -0.08in
    \caption{\label{fig:planes}\textbf{Ten standard planes.}}
    \vspace{-10pt}
\end{figure}
For the probe guidance task, we consider ten target standard planes recommended by the American Society of Echocardiography \cite{mitchell2019guidelines}, as shown in Figure \ref{fig:planes}. These planes include: Parasternal Long-Axis (PLAX), Parasternal Short-Axis Aortic Valve (PSAX-AV), Pulmonic Valve (PSAX-PV), Mitral Valve (PSAX-MV),  Papillary Muscles  (PSAX-PAP), Level of Apex (PSAX-APEX), Apical Four-Chamber (A4C), Apical Five-Chamber (A5C), Apical Three-Chamber (A3C), and Apical Two-Chamber (A2C). Professionals manually annotate the timestamps and frames corresponding to these planes, which serve as the ground truth for the probe guidance task. The dataset is divided into separate training (284 scans) and testing (72 scans) sets, with no overlap of individuals between the two. 



\section{Tasks and Baselines}
The probe guidance task in our study involves predicting the probe's movement toward ten standard planes. The prediction can rely on either a single image or incorporate past visual-motion data. Specifically, in an ultrasound scan comprising $T$ frames, represented as $\{\bm I_t,\bm p_t\}_{t=1}^T$, experts identify the timestamps at which the ten standard planes are observed, denoted as $s_1,s_2,\dots,s_{10}$. For each timestep $t$ with image $\bm I_t$ and corresponding pose $\bm p_t$, the relative pose to the $k$-th standard plane is computed as $\bm a_t^{(k)} = \bm p_{s_k} \cdot \bm p_t^{-1}$. The model’s objective is to predict these movements $\bm a_t$ based on the available visual-motion data. 

The probe pose is represented in six degrees of freedom $\bm p\in \mathbb R^6$, where the first three components represent translations (x,y,z) in millimeters, and the last three correspond to rotations (yaw, pitch, roll) in degrees. For model evaluation, we calculate the mean absolute error separately for translation and rotation components, as detailed in Table 1.

We employ two evaluation protocols (single-frame and sequential) in our study, as described in Section 5.2. Below, we provide a detailed introduction to each protocol.

\subsection{Single-Frame Protocol}
In the single-frame protocol, the model predicts the probe's movement toward all ten standard planes using a single ultrasound image as input. This setup evaluates the representation quality of pre-trained visual models in a cost-efficient manner. Two-layer MLPs are appended to the pre-trained backbones, and the entire model undergoes full fine-tuning. The evaluation metric is computed as the average error across all frames in the test set. To improve evaluation efficiency, the frame rate is reduced to 6 fps.

In this protocol, we evaluate the performance of our visual encoder, pre-trained using world modeling tasks, against a diverse selection of pre-trained models. These include DeiT \cite{touvron2021training}, DINOv2 \cite{oquab2023dinov2}, BioMedCLIP \cite{zhang2023biomedclip}, LVM-Med \cite{mh2024lvm}, US-MoCo \cite{mocov3}, US-MAE \cite{mae}, USFM \cite{jiao2024usfm}, EchoCLIP \cite{christensen2024vision}. For consistency, we use the ViT-Small variant of each method whenever available. Below, we provide an overview of these baselines:
\begin{itemize}
    \item \textbf{DeiT} \cite{touvron2021training} is a family of vision transformers trained on the ImageNet dataset \cite{deng2009imagenet}.
    \item \textbf{DINOv2} \cite{oquab2023dinov2} is a state-of-the-art self-supervised vision foundation model trained on a wide range of general-domain images. The training algorithm mainly follows DINO \cite{dino} and iBOT \cite{zhou2021ibot}.
    \item \textbf{BioMedCLIP} \cite{zhang2023biomedclip} is a multimodal biomedical foundation model pre-trained on 15 million medical image-text pairs using contrastive learning. We utilize only the visual encoder component of this model for our comparisons.
    \item \textbf{LVM-Med} \cite{mh2024lvm} employs a graph-matching formulation for contrastive learning, enabling it to integrate multiple medical imaging modalities, including ultrasound, into a single versatile framework.
    \item \textbf{US-MoCo} \cite{mocov3} is an adaptation of the MoCo framework to our dataset. MoCo employs a momentum encoder to create a dynamic dictionary for stable and effective representation learning. We pre-train a ViT-Small model on our ultrasound dataset using the MoCov3 codebase, training for 150 epochs with a learning rate of $1.5\times 10^{-4}$, weight decay of $0.1$, and batch size of $1024$. 
    \item \textbf{US-MAE} \cite{mae} is an adaptation of the MAE framework to our dataset. MAE is an encoder-decoder framework for mask image modeling. For this adaptation, we train a ViT-Small with a four-layer decoder, a masking ratio of 0.75, over 300 epochs. The training setup includes a learning rate of $6\times 10^{-4}$, weight decay of $0.05$, and batch size of $1024$.
    \item \textbf{USFM} \cite{jiao2024usfm} is an ultrasound-specific vision foundation model trained on over 2 million ultrasound images using a spatial-frequency dual mask modeling approach.
    \item \textbf{EchoCLIP} \cite{christensen2024vision} is a multimodal foundation model for echocardiogram interpretation. The model is trained on more than 1 million ultrasound image-text pairs using contrastive learning. We utilize only the visual encoder component of this model for our comparisons.
\end{itemize}

\subsection{Sequential Protocol}
The sequential protocol simulates a deployment scenario, where the model predicts the probe's movement toward unvisited planes based on past visual-motion data up to the current timestep (visited planes are excluded from the prediction error calculation). It provides a more holistic assessment of probe guidance frameworks. In this setting, we use our pre-trained visual encoder as the backbone for all baselines. Specifically, at the timestep $t$ of a scan, the model uses historical visual-motion data before $t$ to predict the standard planes that are yet to be visited.
The history data $\mathcal H_t$ and the target plane indices $\mathcal K_t$ are defined by:
\begin{equation}
    \begin{aligned}
    \mathcal H_t&=\{(\bm I_{t'},\bm p_{t'})|t'<t\},\\
    \mathcal K_t&=\{k|s_k \ge t\},
    \end{aligned}
\end{equation}
where $s_k$ is the timestep when the $k$-th plane is visited. To construct the model inputs, 
we sample $N$ visual-motion pairs $\{\bm I_{t_i}, \bm p_{t_i}\}_{i=1}^N$ from $\mathcal H_t$ using a decayed density sampling rate. This approach ensures that recent observations are prioritized while retaining a representative selection of past data.
The sampled timesteps $t_i$ are computed as: 
\begin{equation}
    t_i = \operatorname{Round}\left(t + \frac{t}{\alpha N} \log \frac {i}{N}\right),\quad i=1,\dots,N,
\end{equation}
where $\alpha$ is a scaling factor. By default, we sample $N=8$ frames from the history with $\alpha=0.4$. If the history contains fewer than eight frames, we allow repeated sampling to meet the required count.
To ensure a symmetric evaluation, we assess the model in both forward and reverse directions for each scan. In the reverse direction, the scan begins at the last frame. The historical data $\widetilde{\mathcal{H}}_t$ and target plane indices $\widetilde{\mathcal K}_t$ at timestep $t$ are defined as:
\begin{equation}
    \begin{aligned}
    \widetilde{\mathcal H}_t&=\{(\bm I_{t'},\bm p_{t'})|t'\ge t\},\\
    \widetilde{\mathcal K}_t&=\{k|s_k < t\}.
    \end{aligned}
\end{equation}
The final error metric is averaged over both forward and reverse directions and all timesteps across the scans. For computational efficiency, the frame rate is reduced to 3 fps during this evaluation.

In this protocol, we evaluate the complete \methodname{} framework, which incorporates the proposed motion-aware attention mechanism, by comparing it against existing probe guidance frameworks. These include US-GuideNet \cite{droste2020automatic}, Decision-Transformer \cite{chen2021decision}, and Sequence-aware Pre-training \cite{jiang2024sequence}. To ensure a fair comparison and isolate the impact of our motion-aware modeling, all baselines use the same visual encoder. The visual encoder extracts average-pooled image features, which are subsequently passed to the respective probe guidance frameworks. Below, we provide detailed descriptions of these baselines:
\begin{itemize}
    \item \textbf{US-GuideNet} \cite{droste2020automatic} is originally designed for freehand obstetric ultrasound probe guidance.
    In our implementation, we adopt its model design, which processes sequential inputs in the form:
    \begin{equation}
    \label{eq:seq}
        \{\bm I_1,\bm p_{1\rightarrow 2},\bm I_2,\bm p_{2\rightarrow 3}, \bm I_3,\dots, \bm I_N\}.
    \end{equation}
    Here, $\bm p_{i\rightarrow i+1}$ denotes probe movements between consecutive frames.
    Visual and motion features are projected and concatenated before being aggregated using a gated recurrent unit (GRU).
    \item \textbf{Decision-Transformer} \cite{chen2021decision} models trajectories within a Markov Decision Process using a causal transformer. For our task, we adapt this architecture by feeding interleaved states (images) and actions (probe movements) using the same input structure as Equation \eqref{eq:seq}. The interleaved sequence is passed through a two-layer causal transformer, with the output of the final token feeding into a guidance prediction head for downstream tasks.
    \item \textbf{Sequence-aware Pre-training} \cite{jiang2024sequence} utilizes a bidirectional transformer to process interleaved visual-motion sequences, adhering to the same input format as Equation \eqref{eq:seq}. The model is pre-trained using a visual-motion mask modeling strategy to enhance historical data aggregation. During fine-tuning, an extra mask token is appended to the sequence for probe movement prediction. 
\end{itemize}

\begin{algorithm}[tb]
   \caption{PyTorch-style pseudocode for motion-aware attention.}
   \label{alg:motion_attn}
   
    \definecolor{codeblue}{rgb}{0.25,0.5,0.5}
    \lstset{
      basicstyle=\fontsize{7.2pt}{7.2pt}\ttfamily\bfseries,
      commentstyle=\fontsize{7.2pt}{7.2pt}\color{codeblue},
      keywordstyle=\fontsize{7.2pt}{7.2pt}\color{BrickRed},
    }
\begin{lstlisting}[language=python]
# B: batch size
# N: number of frames
# D_img: dimensionality of the image features
# D_mo: dimensionality of the motion features
# D: dimensionality of attention features
# x_img:  image features  shaped (B, N, D_img)
# x_motion:  motion features shaped (B, N, N, D_mo)

def motion_aware_attn(x_img, x_motion):
    # expand image features
    x_img_exp = x_img.unsqueeze(1).expand(B,N,N,D_img)

    # compute query, key, and value
    Q = mlp_q(x_img) # BxNxD
    K = mlp_k(concat(x_img_exp, x_motion)) # BxNxNxD
    V = mlp_v(concat(x_img_exp, x_motion)) # BxNxNxD

    # perform attention
    logits = einsum('bid,bijd->bij', Q,K) / (D ** 0.5)
    attn = softmax(logits, dim=-1)
    return einsum('bij,bijd->bid', attn, V)

\end{lstlisting}
\end{algorithm}

\section{Implementation Details}
\subsection{Pre-training}
\textbf{Architecture and optimization.}
\methodname{} is pre-trained from scratch by jointly performing spatial and motion world modeling. The context encoder is a ViT-S/16, while the target encoder is an exponential moving average (EMA) of the context encoder with a starting decay rate of 0.996, which gradually increases to 1.0 following a cosine schedule.
 The predictor is a 6-layer transformer with a width of 384. Input images are resized to $224\times 224$. The model is optimized using the AdamW optimizer with $\beta_1=0.9$ and $\beta_2=0.999$, an initial learning rate of $10^{-3}$, and a weight decay of 0.05. Training spans 300 epochs, with a 40-epoch linear warm-up followed by cosine decay. The default batch size is 1024, and training takes approximately 14 hours on four A100 GPUs.

\textbf{Spatial world modeling.} Following \cite{ijepa}, the context image is masked using four rectangular blocks with scales ranging from $(0.15, 0.2)$. The visible regions are further reduced by up to 15\%, increasing the task's difficulty. Only visible patches are processed by the context encoder, whereas the target encoder takes the entire image as input. In the predictor, mask tokens, enriched with positional encodings corresponding to the masked patches, are concatenated with context tokens. A smoothed L1 loss is computed between the predicted and target outputs at the masked locations.

\textbf{Motion world modeling.} We randomly sample two frames $\bm I_a,\bm I_b$ along with their respective poses $\bm p_a,\bm p_b$ from a scan and compute their relative pose difference $\bm p_{a\rightarrow b}=\bm p_b \cdot \bm p_a^{-1}$. Frame $\bm I_a$ is used as input to the context encoder, while frame $\bm I_b$ serves as the target. The motion encoder $A_\psi$ is a two-layer MLP with a hidden dimension of 384, producing motion feature $\bm z_{a\rightarrow b}=A_\psi(\bm p_{a\rightarrow b})$.  These features are embedded into a mask token and concatenated with context tokens before being passed to the predictor. The predictor generates $\hat{h}_y$, a prediction of the average-pooled target feature $h_y$.  Before computing the InfoNCE loss, $\hat{h}_y$ and $h_y$ are projected using projectors $P$ and $P'$, where $P'$ is an EMA of $P$. For simplicity, we skip the projector in Equation (4) of the main paper. The loss, including the projector, is defined as:
\begin{equation}
\begin{aligned}
    &\quad \mathcal L_{motion} \\
    &= -\frac 1 B \sum_{i=1}^B\log \frac{\exp(P(\hat{h}_{y_i})^\top \cdot P'(h_{y_i})/\tau)}{\sum_j \exp(P(\hat{h}_{y_i})^\top \cdot P'(h_{y_j})/\tau)},
\end{aligned}
\end{equation}
where $B$ is the batch size and $\tau$ is the temperature (set to $0.1$ by default). The loss can be symmetrized by swapping the context and target roles.

\textbf{Joint modeling.} The integration of spatial and motion world modeling follows a unified pipeline. Specifically, for the frames $\bm{I}_a$ and $\bm{I}_b$ used in motion modeling, some regions in the context frame $\bm{I}_a$ are masked. The predictor simultaneously performs two tasks: (1) reconstructing masked regions in the context frame and (2) predicting features of the target frame based on motion information. The predictions and targets for these tasks are defined as:
\begin{equation}
\begin{aligned}
    h_x &= f_\theta(\operatorname{Mask}(\bm I_a, M)),\\
    \hat{h}_y^{\operatorname{spatial}} &= g_\phi(h_x+p_x; \{m+ \operatorname{PE}(c)\}_{c\in M}), \\
    h_y^{\operatorname{spatial}} &= \{f'_{\theta'}(\bm I_a)_c\}_{c\in M}, \\
    \hat{h}_y^{\text{motion}} &= g_\phi(h_x; m+\bm z_{a\rightarrow b}),\\
    h_y^{\operatorname{motion}} &= \operatorname{AvgPool}(f'_{\theta'}(\bm I_b)),
\end{aligned}
\end{equation}
where $\hat{h}_y^{\operatorname{spatial}},h_y^{\operatorname{spatial}}$ are prediction and target for spatial modeling, and $\hat{h}_y^{\operatorname{motion}},h_y^{\operatorname{motion}}$ are for motion modeling. 
The total loss combines both objectives: $    \mathcal L_{\text{total}} = \mathcal L_{\text{spatial}} + \lambda \mathcal L_{\text{motion}}$, where $\lambda=0.1$ balances the scale of the two losses.

\subsection{Fine-tuning}

\textbf{Motion-aware attention.}
Algorithm \ref{alg:motion_attn} provides the pseudocode for the proposed motion-aware attention mechanism. The pre-trained visual encoder $f_\theta$ and motion encoder $A_\psi$ extract visual and motion features, $\bm{h}_i$ and $\bm{z}_{i \rightarrow j}$, for frames $i, j \in [1, N]$. Two MLPs process their concatenation to generate keys $K_j^{(i)}$ and values $V_j^{(i)}$ as follows:
\begin{equation}
    K_j^{(i)} = \operatorname{MLP}_{k}(\bm h_j, \bm z_{i\rightarrow j}),\ V_j^{(i)} = \operatorname{MLP}_{v}(\bm h_j, \bm z_{i\rightarrow j}).
\end{equation}
Queries are derived from the image features using another MLP: $Q_i = \operatorname{MLP}_q(\bm{h}_i)$. The model applies scaled dot-product attention with four attention heads and a hidden dimension of 384. The resulting attention outputs are passed through ten independent MLPs to predict probe movements to ten standard planes relative to the current pose.

\textbf{Optimization.} The model is optimized using AdamW with $\beta_1 = 0.9$, $\beta_2 = 0.999$, and an initial learning rate of $1 \times 10^{-4}$. Training uses 15,000 iterations with a batch size of 256 for single-frame and 64 for sequential protocols. Additional settings include weight decay of 0.05, drop path of 0.1, layer-wise learning rate decay of 0.65, and random brightness/contrast augmentations.

\subsection{Visualizations}

\textbf{World model predictor outputs (Figure 7).} To better understand the predictor outputs of our world model, we train a diffusion model to reconstruct target pixel values conditioned on the representation $\hat{h}_y$ produced by the predictor. This guidance representation is first projected to a 512-dimensional vector, which is then integrated into the diffusion model via conditional batch normalization layers \cite{dumoulin2016learned}. For spatial world modeling, the diffusion model is conditioned on the average-pooled predictor outputs corresponding to the masked regions. For motion world modeling, the diffusion model takes the predictor output vector as its conditioning signal. We train separate diffusion models for the two world modeling tasks, with both models trained for 300,000 iterations and generating images at resolution $128\times 128$.

\textbf{Analysis of attention scores (Figure 8). } We evaluate the proposed motion-aware attention mechanism by visualizing attention scores across a set of eight visual-motion pairs. Some of these pairs include noisy frames with minimal usable information. For this input, we visualize the $8 \times 8$ attention score matrices across all four attention heads. Each matrix entry, located at the $i$-th row and $j$-th column, represents the attention score of query $i$ attending to key $j$.

\textbf{Analysis of plane features (Figure 9). } We extract average-pooled representations of all standard plane images identified by professionals in the training set. These representations are visualized in a 2D space using t-SNE. The visualization highlights how well the model clusters images of similar planes.

\end{document}